\documentclass[conference]{IEEEtran}
\IEEEoverridecommandlockouts

\usepackage{amsmath,amssymb,amsfonts}
\usepackage{graphicx}
\usepackage{url}
\usepackage{cite}
\usepackage{xcolor}
\usepackage{booktabs}
\usepackage{subcaption}
\usepackage{cuted}

\begin{document}
\title{SAGE: Ergodic Control for Autonomous and Adaptive Inspection of Subsea Infrastructure}

\renewcommand{\thefootnote}{\arabic{footnote}}
\setcounter{footnote}{0}

\author{Markus Buchholz\textsuperscript{1}\thanks{Norwegian Defence Research Establishment (FFI), Kjeller, Norway. \texttt{markus.buchholz@ffi.no}}%
\and Ignacio Carlucho\textsuperscript{2}\thanks{School of Engineering \& Physical Sciences, Heriot-Watt University, Edinburgh, UK. \texttt{\{ignacio.carlucho, y.r.petillot\}@hw.ac.uk}}%
\and Yvan R. Petillot\textsuperscript{2}}

\maketitle

\begin{abstract}

Subsea Christmas Trees (XTs) are underwater structures that use valves for directing oil flow, needing constant inspection. But not every valve carries the same risk at the same time: a valve with a suspected leak needs to be revisited far more often than one with a clean history, and that risk picture changes during the mission as new leaks are found. To handle this, we present SAGE (Semantic and Adaptive Generative Ergodicity), an ergodic-control architecture that allocates vehicle time in proportion to a live, sensor-derived risk distribution rather than a scripted route. We study a two-XT scenario, with five valves in total, and compare a fixed-loop A* tour against SAGE.
Both methods can be tuned to spend similar total time near a high-risk valve, but only ergodic control also checks it more often: in simulation, a dominant-risk valve was revisited every 5.8 s under ergodic control against a fixed 8.1 s for every valve under A*, regardless of risk, so a leak can go unnoticed for barely two-thirds as long.
Because the tracked distribution is recomputed rather than planned once, a newly detected leak shifts vehicle behavior on the next control cycle with no explicit re-planning step and no operator in the loop, which a fixed tour cannot do without a discrete re-route. We derive the ergodic control law behind this behavior and report simulation results on the five-valve scenario.
\end{abstract}

\section{Introduction}
Inspecting subsea production infrastructure means repeatedly checking a small set of known-risk points, valves, flanges, and connectors, rather than searching an unknown space. Today this is carried out largely by divers and remotely operated vehicles (ROVs) under direct human supervision \cite{nauert2023inspection}, following fixed schedules rather than responding to risk as it emerges. The difficulty is dividing attention among them when their relative risk is uneven and non-stationary: a leak detected mid-mission should immediately pull more of the vehicle's time, without a human re-planning the route. A fixed inspection loop treats every point the same, and making it risk-aware normally means adding a rule that turns severity into a wait time at each point, then redoing the route by hand whenever risk changes.
Ergodic control addresses this problem directly. Instead of planning a route, it acts as an \textbf{autonomous} coverage controller. It aligns the time-averaged statistics of the trajectory with a target risk distribution. This means that revisit frequency, not just time spent, measures importance automatically and adjusts within a control cycle. \textbf{No human is involved in this process.}

\section{Scenario and Ergodic Control}
\label{sec:scenario}
\begin{figure*}[t]
\centering
\includegraphics[width=0.9\textwidth, keepaspectratio]{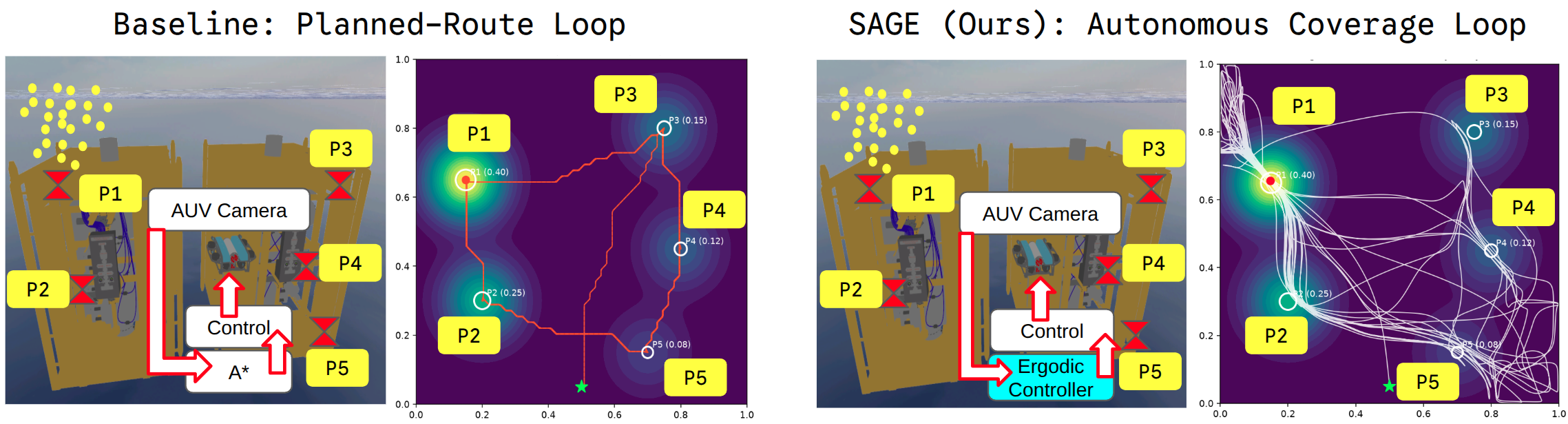}
\caption{Valve P1 leak response during BlueROV2 inspection of two subsea XTs in Gazebo \cite{buchholz2025tethered}. \textbf{Left (A*):} The leak spikes risk at P1, but adapting requires stopping to manually re-plan and re-issue routes; all valves remain at a fixed 8.1~s visit interval. \textbf{Right (SAGE, ours):} The risk field updates dynamically without re-planning, automatically re-allocating coverage on the next cycle to revisit P1 every 5.8~s (about $1.4\times$ more often than A*'s fixed 8.1~s) fully autonomously.}
\label{fig:loops}
\end{figure*}

Fig.~\ref{fig:loops} shows the target scenario: two subsea XTs, one with two monitored valves and the other with three, five points of interest in total, non-collinear and unevenly spaced. Each valve $i$ carries a time-varying risk weight $w_i(t)$; in the full SAGE architecture (Fig.~\ref{fig:loops}, right) this weight updates autonomously from operator intent, from a camera or sonar flagging a new leak, or from both fused together. For the benchmark below it is instead hand-set to fixed values per scenario, so the coverage behaviour can be validated against a known risk profile before closing the perception loop.

Unlike a path planner, the controller never represents a route: it turns the current coverage error into a velocity every cycle, so the trajectory emerges rather than being planned. Let $X\subset\mathbb{R}^2$ be the normalized workspace and $\phi(x,t)$ the current risk over that workspace, built from a weighted mixture of the five valve positions,
\begin{equation}
\phi(x,t) = \frac{1}{Z}\sum_{i} w_i(t)\,\mathcal{N}\big(x;\,\mu_i,\,\Sigma_i\big),
\end{equation}
with $\mathcal{N}(x;\mu_i,\Sigma_i)$ a Gaussian density, so valve $i$ contributes a bump at its known position $\mu_i$, of spread $\Sigma_i$, scaled by its live risk weight $w_i(t)$; $Z$ rescales the sum so $\phi$ integrates to one. A trajectory is ergodic with respect to $\phi$ when the fraction of time it spends in any region converges to the risk mass $\phi$ assigns there \cite{mathew2011metrics}, so a valve carrying twice the risk should, over time, receive roughly twice the visits.

Comparing a distribution against a trajectory directly is awkward, so both are projected onto the same Fourier basis $F_k(x)=\tfrac{1}{h_k}\prod_j\cos(k_j\pi x_j/L_j)$: $\phi_k=\int_X\phi(x)F_k(x)\,dx$ gives the target's spectral coefficients, and $c_k(t)=\tfrac{1}{t}\int_0^t F_k(x(\tau))\,d\tau$ gives the same for the trajectory's own running time-average. Their weighted mismatch, emphasizing the coarse shape of $\phi$ over its fine detail, is exactly the ergodic metric of Mathew and Mezi\'{c} \cite{mathew2011metrics}. Rather than minimizing that metric over the whole mission, the spectral multiscale coverage law of Mathew and Mezi\'{c} \cite{mathew2011metrics}, which we implement on top of the reference ergodic-control-sandbox \cite{murpheylab_sandbox}, drives it down one instant at a time: the vehicle moves along $S(x,t)$ at constant transit speed, where
\begin{equation}
\begin{aligned}
S(x,t) &= -\sum_k \Lambda_k\big(c_k(t)-\phi_k\big)\,\nabla_x F_k(x), \\
\Lambda_k &= (1+\|k\|^2)^{-\frac{n+1}{2}},
\end{aligned}
\end{equation}
downweights high frequencies so coarse coverage is corrected before fine detail. Because $\phi$ enters $S$ only through $\phi_k$, raising any $w_i(t)$ changes the commanded direction on the very next control cycle. This is what lets a newly flagged leak redirect the vehicle immediately: reflecting the same event in an A* tour instead means detecting it, recomputing wait times, and re-issuing a route, a discrete step that normally waits on an operator or supervisory trigger, whereas here coverage, prioritization, and reprioritization all live inside one continuous control law with no planning step anywhere.

\section{Results and Discussion}
\label{sec:results}
The baseline is a single closed-loop A* tour visiting all five valves once per lap on an 8-connected grid (shortest-tour order), with per-valve wait time set proportional to $w_i$ so that A* can match ergodic control on time spent per valve; its revisit interval, however, is fixed by lap geometry at 8.1~s regardless of weight, since the loop itself encodes no priority. We ran the five-valve layout of Fig.~\ref{fig:loops} for 300~s under three hand-set risk profiles: equal weights (20\% each, a sanity check), graded weights (40/25/15/12/8\%), and the one-extreme profile of Table~\ref{tab:results} (55/20/10/8/7\%), representing a single confirmed leak at valve P1 against four routine valves.

\begin{table}[t]
\centering
\small
\caption{One-extreme profile (leak at valve P1), per-valve outcome, A* vs. ergodic control (EC), 300~s mission.}
\label{tab:results}
\renewcommand{\arraystretch}{0.92}
\begin{tabular}{lccccc}
\toprule
\textbf{Valve} & \textbf{Weight} & \multicolumn{2}{c}{\textbf{A*}} & \multicolumn{2}{c}{\textbf{EC}} \\
 & & visits & interval & visits & interval \\
\midrule
P1 (leak) & 0.55 & 37 & 8.1~s & 50 & 5.8~s \\
P2 & 0.20 & 37 & 8.1~s & 10 & 26.8~s \\
P3 & 0.10 & 37 & 8.1~s & 0  & --- \\
P4 & 0.08 & 37 & 8.1~s & 7  & 43.5~s \\
P5 & 0.07 & 37 & 8.1~s & 15 & 18.4~s \\
\bottomrule
\end{tabular}
\end{table}

Table~\ref{tab:results} presents the results. A* revisits every valve 37 times over the mission regardless of risk, while ergodic control tracks the same weights, checking leaking valve P1 50 times (every 5.8~s) against zero visits at P3. Correlating revisit count with weight is undefined for A* (zero variance across valves) but $+0.88$ (graded) and $+0.93$ (one-extreme) for ergodic control. P3's zero visits are a real limitation. 
Unconstrained ergodic control only bounds the long-run average revisit rate, not any single point's rate within a finite mission, so a strongly peaked risk distribution can starve the lowest-weight valves. The realized trajectories are shown in Fig.~\ref{fig:loops} for the graded-weight case; the brief corner excursions visible in the ergodic panel are a small Fourier-basis artifact (2--5\% of mission time), not a bug, and do not affect the numbers in Table~\ref{tab:results}.

The lack of visits to P3 in Table~\ref{tab:results} is a real limitation. One way to solve this would be to inc a minimum weight floor ($w_i(t)\ge\epsilon>0$) or a hard revisit-interval constraint layered on the ergodic objective would guarantee every valve is checked at least once per mission, and is a natural next addition to the control law. Beyond that, the next steps are to test a broader set of scenarios and XT layouts, close the perception loop by driving $w_i(t)$ directly from real-time camera-based leak detection instead of the hand-set profiles used here, and validate all of this in real-world trials on a physical vehicle, including a mid-mission leak-injection experiment. 

A more complete version of SAGE would integrate detection confidence into $w_i(t)$ using a Bayesian update and determine operator intent with a language model, as shown in Fig. 1 (right). Since the controller only requires a single risk weight per point of interest rather than specific geometry for a task, the same structure can apply beyond valve inspection. It can be used for other coverage tasks determined by a risk or interest field, such as search-and-rescue casualty localization, monitoring environmental hazards like harmful algal blooms or oil spills, or litter collection by an autonomous surface vehicle (ASV) in fjords and coastal waters.

{\footnotesize
\bibliographystyle{IEEEtran}
\bibliography{references}
}

\end{document}